\documentclass[letterpaper, 10 pt, conference]{ieeeconf}  

\IEEEoverridecommandlockouts                              

\usepackage{amsmath} 
\usepackage{amssymb}  
\usepackage{lettrine}
\usepackage{kotex}
\usepackage{graphicx}
\usepackage{algorithm}
\usepackage{algpseudocode}
\usepackage{multirow}
\usepackage{booktabs}
\usepackage{adjustbox}
\usepackage{subfigure}
\usepackage{hyperref}

\title{\LARGE \bf
Learning Multi-Task Transferable Rewards via Variational Inverse Reinforcement Learning
}

\author{Se-Wook Yoo,  Seung-Woo Seo 
    \thanks{ The authors are with the Department of Electrical and Computer Engineering, Seoul National University, Republic of Korea.
        {\tt\small tpdnr1360@snu.ac.kr, sseo@snu.ac.kr}}
    }

\begin{document}

\maketitle
\thispagestyle{empty}
\pagestyle{empty}

\begin{abstract}

Many robotic tasks are composed of a lot of temporally correlated sub-tasks in a highly complex environment. It is important to discover situational intentions and proper actions by deliberating on temporal abstractions to solve problems effectively. To understand the intention separated from changing task dynamics, we extend an empowerment-based regularization technique to situations with multiple tasks based on the framework of a generative adversarial network. Under the multitask environments with unknown dynamics, we focus on learning a reward and policy from the unlabeled expert examples. In this study, we define situational empowerment as the maximum of mutual information representing how an action conditioned on both a certain state and sub-task affects the future. Our proposed method derives the variational lower bound of the situational mutual information to optimize it. We simultaneously learn the transferable multi-task reward function and policy by adding an induced term to the objective function. By doing so, the multi-task reward function helps to learn a robust policy for environmental change. We validate the advantages of our approach on multi-task learning and multi-task transfer learning. We demonstrate our proposed method has the robustness of both randomness and changing task dynamics. Finally, we prove that our method has significantly better performance and data efficiency than existing imitation learning methods on various benchmarks.

\end{abstract}

\section{INTRODUCTION}

\begin{figure}[t]
    \centering
    \subfigure[Mutual Information]{
        \includegraphics[width=.46\columnwidth]{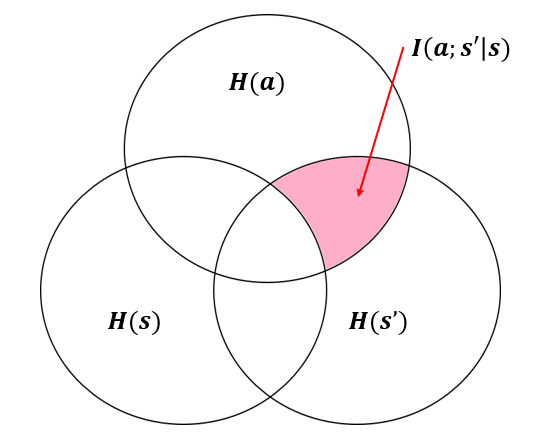}
        }
    \subfigure[Situational Mutual Information]{
        \includegraphics[width=.46\columnwidth]{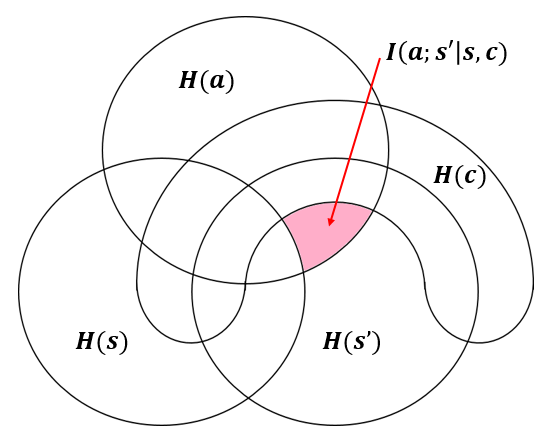}
        }
    \caption{ \textbf{Left:} Information diagram for existing empowerment-based regularization technique. The variables are the current state $s$, action $a$, and next state $s'$. They use mutual information $I(a ; s' \vert s)$ (pink region on (a)) as an internal reward \textbf{Right:} We expand mutual information (a) to situational mutual information $I(a ; s' \vert s, c)$ (b) by introducing a sub-task variable $c$.}
    \vspace{-0.5cm}
    \label{fig:information_diagram}
\end{figure}

\lettrine{R}{ecently}, a paradigm of learning from demonstration (LfD) \cite{argall2009survey} that learns various skills by utilizing expert demonstrations as a teacher has been developed. Imitation learning (IL) \cite{pomerleau1998autonomous}, a branch of LfD, focuses on directly learning a policy that mimics the teacher behaviors. Behavior cloning (BC) \cite{pomerleau1991efficient}, which simply imitates the policy through supervised learning, has difficulty in solving high-dimensional problems due to compounding errors caused by distribution shift. Although many variants \cite{fu2017learning, li2017infogail} of generative adversarial imitation learning (GAIL) \cite{ho2016generative} have successfully resolved simple primitive tasks, there are a variety of hindrances to achieving a breakthrough in complex multitask environments. Most applications require the setting of sub-tasks and the selection of the appropriate actions according to a predefined hierarchical structure. It is not only difficult to design a general hierarchical relationship with predefined rules, but it is also wasteful to label sub-tasks. Although Directed-Info GAIL (DIGAIL) \cite{sharma2018directed} tried to find a hierarchical policy in an unsupervised way, the learned policy did not adapt to cases different from those seen in the demonstration. This is because the discriminator simply compares trajectories sampled from demonstrations with the generated trajectories. To overcome this limitation, adversarial inverse reinforcement learning (AIRL) \cite{fu2017learning} has been proposed. It can restore a robust state-only dependent reward function by adding a potential-based shaping term inspired by \cite{ng1999policy}. To reconstruct state-action dependent rewards, Empowered AIRL (EAIRL) \cite{qureshi2018adversarial} 
learns a potential function with empowerment (i.e., a theoretical measure that maximizes mutual information in Fig. \ref{fig:information_diagram}(a)). Even though the policy can be adapted to transformed dynamics, it can only handle a single task. Unlike previous works, we focus on the multi-task transfer learning problem and suggest a solution by extending previous works and unraveling the connection. 

To infer the sub-task in an unsupervised manner, we adopt the architecture of DIGAIL developed from the options framework \cite{sutton1999between} using the supervision acquired from unsegmented demonstrations. However, DIGAIL has only resolved a few simple tasks under known dynamics. For example, when the order of sub-tasks is changed, it is difficult for the prior method to handle the problem. To overcome this limitation, we consider the relation of the current state, action, sub-task, and next state, as shown in Fig. \ref{fig:information_diagram}(b). To learn a reward function disentangled from task dynamics, we modify empowerment as the maximum of situational mutual information called situational empowerment by introducing a sub-task variable. Subsequently, we learn situational empowerment via maximization of the variational lower bound of the situational mutual information. It prevents the policy from overfitting into expert demonstrations when the relation of task transition is changed. The tractable optimization for our proposed method, Situational EAIRL (SEAIRL), is explained in Section \uppercase\expandafter{\romannumeral4}. We validate the robustness of our model to environments with randomness and multiple scenarios in Section \uppercase\expandafter{\romannumeral5}. 

As mentioned above, the causal confusion of tasks and behavior in a certain state hinders the learning of the multi-task reward function. The proposed method for solving the aforementioned problems presents the following three main contributions. The first is to successfully learn a potential expression similar to a human's temporal abstraction in an unsupervised manner to increase the interpretability of deep learning. Second, by extending the existing empowerment-based regularization technique, our method successfully restores the robust reward function conditioned on a sub-task separated from the dynamic environment. Finally, we reduce sampling complexity and show state-of-the-art performance that exceeds previous baselines in various environmental settings, which are composed of a few simple tasks, complex multi-tasks, and transferable multi-tasks.

\section{Related Work}

\subsection{Imitation Learning}

IL aimed to directly learn policy $\pi$ that could mimic expert behaviors $\pi_{E}$ from expert trajectories $\tau_{E}$. To mitigate compounding errors, GAIL \cite{ho2016generative} approaches the imitation learning problem as an adversarial learning framework. An agent's policy $\pi$ serves as a generator while the discriminator $D$ represents a local reward function that differentiates samples from expert policy $\pi_E$ and the ones generated from policy $\pi$. The objective is given in Eq. \ref{eq:gailobjective}.

\begin{equation}
    \resizebox{0.9\columnwidth}{!}{$
        \underset{\pi}{\min} \, \underset{D}{\max} \, \mathbb{E}_{\pi}[ \log D(s,a)] + \mathbb{E}_{\pi_{E}} [1 - \log D(s,a)] - \lambda H(\pi) 
    $}
    \label{eq:gailobjective}
\end{equation}

To distinguish different types of behaviors in expert demonstration $\tau_{E}$, InfoGAIL \cite{li2017infogail} introduced a latent variable $c$ into the existing policy $\pi$. InfoGAN \cite{chen2016infogan} mentioned that high mutual information $I(c ; \tau)$ incentivized $\pi$ to use $c$. Inspired by InfoGAN, InfoGAIL induced a variational lower bound $L_{q}(\pi, Q)$ of the mutual information $I(c ; \tau)$ and then added it to the loss function in GAIL. To reduce the dependency of the trajectory from the entire to the current time, DIGAIL \cite{sharma2018directed} modified $L_{q}(\pi, Q)$ by replacing  $I(c ; \tau)$ with directed or causal information flow $I(\tau \rightarrow c)$. This yielded the following lower bound in Eq. \ref{eq:lowerboundposterior} where $\tau_{1:t}$ is $(s_{1},\cdots, a_{t-1},s_{t})$, $H(c)$ is entropy of the posterior $Q$ and the prior distribution $p(c_{1:t})$ is pre-trained from the variational auto-encoder (VAE) \cite{kingma2013auto}.

\begin{align}
    &\sum_{t} \mathbb{E}_{p(c_{1:t}),
    \pi(a_{t-1} \vert s_{t-1}, c_{1:t-1})} [\log Q(c_{t}\vert c_{1:t-1}, \tau_{1:t})] + H(c) \nonumber \\
    & = L_{q} (\pi, Q) \,\leq\, I (\tau \rightarrow c) 
    \label{eq:lowerboundposterior}
\end{align}

Unlike previous works, we aim to recover a portable or transferable reward function that depends on the latent variable $c$. In Section \uppercase\expandafter{\romannumeral4}-B, we explain how to incorporate the above concept into the multi-task IRL framework.

\subsection{Variational Information Maximization}

AIRL \cite{fu2017learning} modeled the expert trajectory distribution $p(\tau_{E})$ with an energy-based model (EBM) $e^{\sum_{t=0}^{T}r(s_{t},a_{t})}$ where the energy function corresponds to the reward function, connected to the sampling-based MaxEnt-IRL framework \cite{finn2016connection}. It can restore the disentangled reward function from dynamics by adding shaping term, such as $f(s,a) = r(s,a) + \gamma \Phi(s') - \Phi(s)$, representing an optimal discriminator with $\frac{e^{f(s,a)}}{e^{f(s,a)} + \pi(a|s)}$. Nevertheless, AIRL can only recover the state-dependent reward function because the irregular features of the action prevent the function from being learned. To solve this limitation, EAIRL uses the empowered reward $\Phi(s)$ that maximizes mutual information $I(a; s' \vert s)$ as the internal reward and inverse model $\Omega(a \vert s, s')$ to normalize the policy. This gives the following lower bound, denoted as Eq. \ref{eq:empoweredlowerbound}, where $w(a \vert s)$ denotes the normalized policy.

\begin{equation}
    \resizebox{0.9\columnwidth}{!}{$
        H(w(a|s)) + \mathbb{E}_{w,P}[\log \Omega(a|s,s')] = L_{I}(w, \Omega) \leq I(a;s'|s)
    $}
    \label{eq:empoweredlowerbound}
\end{equation}

The lower bound $L_{I}(w, \Omega)$ is optimized using the expectation-maximization (EM) algorithm over the distribution of the action and the inverse model. The inverse model $\Omega$ parameterized by $\phi$ minimizes the mean squared error (MSE) between actions $a$ conducted by policy $\pi$ and predicted action $\hat{a}$ from the inverse model $\Omega$, which is based on a supervised maximum log-likelihood problem formulated as Eq. \ref{eq:inversemodelobjective}.

\begin{equation}
    l_{q}(s,a,s') = \vert\Omega_{\phi}(\hat{a}|s,s') - a\vert^{2} 
    \label{eq:inversemodelobjective} 
\end{equation}

$w(a)$ is calculated from the solution of the Lagrange dual problem over $\frac{\partial \hat{I}^{w}}{\partial w} = 0 \; s.t \sum_{a} w(a|s) = 1$. The analytical solution is $w^{*}(a|s) = \frac{e^{u(s,a)}}{Z(s)}$ where $u(s,a) = \beta \mathbb{E}_{P}[ \log \Omega(a|s,s')]$ and $ Z(s) = \sum_{a} u(s,a)$. The normalization term $\log Z(s)$ is equivalent to the empowerment-based potential function $\Phi(s)$ parameterized by $\varphi$. While parameterizing the policy $\pi$ with $\theta$, the MSE between the approximated $w_{\theta, \varphi}(a \vert s) \approx \log \pi_{\theta} (a \vert s) + \Phi_{\varphi}(s)$ and $\log \Omega(a \vert s, s')$ is minimized using Eq. \ref{eq:empowermentobjective}. 

\begin{equation}
    l_{I}(s,a,s') = \vert\log \Omega(a|s,s') - \{ \log \pi_{\theta}(a|s) + \Phi_{\varphi}(s) \}\vert^{2}
    \label{eq:empowermentobjective}    
\end{equation}

To deal with the multi-task problem that is resilient to task transition, our method reformulates the empowerment-based regularization by adding the latent variable $c$ into the above equations, which is explained in Section \uppercase\expandafter{\romannumeral4}-A. 

\subsection{Hindsight Inference}

To overcome limitation of MaxEnt-RL \cite{haarnoja2017reinforcement} or MaxEnt-IRL \cite{ziebart2008maximum} approaches that focus on single primitive tasks, hindsight inference for policy improvement (HIPI) \cite{eysenbach2020rewriting} extended the policy $\pi(a\vert s)$ and reward $r(s,a)$ into task-conditioned policy $\pi(a\vert s, c)$ and multi-task reward $r(s,a, c)$ by introducing a latent variable $c$. HIPI revealed that the process of relabeling sub-tasks with a relabeling distribution $Q(c \vert s, a)$ corresponded to maximizing the multi-task objective, as shown in Eq. \ref{eq:mutitaskobjective}. Furthermore, the objective was optimized at the same time using MaxEnt-RL and MaxEnt-IRL frameworks where $q(\tau)$ is a distribution over previously observed trajectories.

\begin{align}
    &-D_{KL}(q(\tau, c \vert s, a) \;\vert \vert \; p(\tau, c, s, a))
    \nonumber \\
    =&\mathbb{E}_{\substack{c \sim q(c \vert \tau)) \\ \tau \sim q(\tau)}}\Bigg[\sum_{t} r(s,a, c) - \log \pi(a \vert s,c) \nonumber \\ 
    & - D_{KL}(Q(c \vert s,a) \;\vert \vert \; p(c)) - \log Z(c)\Bigg]
    \label{eq:mutitaskobjective}
\end{align}

The expanded soft Q-function is represented by the latent variable $c$ at the first and second terms. An optimal relabeling distribution was found to be the exponential family of the combination of the soft Q-function and a partition function $Z(c)$ that normalizes rewards with different scales. In Section \uppercase\expandafter{\romannumeral4}-B, we discover that our proposed method coincides with Eq. \ref{eq:mutitaskobjective}.

\begin{figure}[t]
    \centering
    \vspace{0.3cm}
    \includegraphics[angle=0, width=\columnwidth]{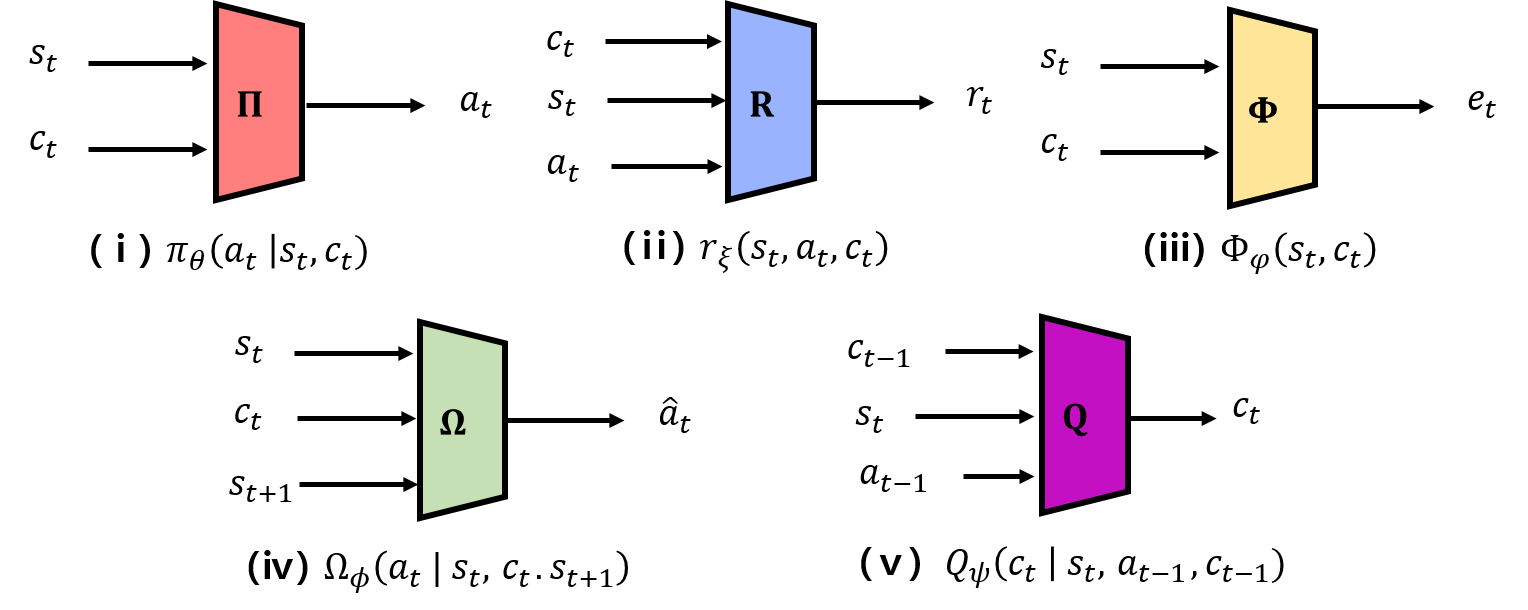}
    \caption{The structures of five networks. We train five models denoted by a policy $\pi$ (red), a reward $r$ (blue), an empowerment-based potential $\Phi$ (yellow), an inverse model $\Omega$ (green) and a posterior $Q$ (purple) with each parameter denoted by $\theta, \xi, \varphi, \phi$ and $\psi$. The above figure shows the input and output information of each network.}
    \vspace{-0.5cm}
    \label{fig:networks}
\end{figure}

\section{Problem Definition}

We analyze the problem under the markov decision process (MDP) expressed as tuple of $(\mathcal{S}, \mathcal{A}, \mathcal{P},\gamma,\mathcal{R})$ where $\mathcal{S}$ denotes state-space, $\mathcal{A}$ means action-space, $\mathcal{P} : \mathcal{S} \times \mathcal{A} \times \mathcal{S} \rightarrow [0,1]$ represents state transition probability distribution and $\gamma \in (0,1)$ is a discount factor. We introduce sub-task-space $\mathcal{C}$ into MDP to handle a multi-task reward function and policy. To be specific, $\mathcal{R}: \mathcal{S} \times \mathcal{A} \times \mathcal{C} \rightarrow \mathbb{R}$ corresponds to the reward function and $\pi:\mathcal{S} \times \mathcal{C} \times \mathcal{A} \rightarrow$ [0,1] is stochastic policy. We focus on solving multi-task learning problem without predefined task-specific knowledge and multi-task transfer learning problem where there are situations that have not been seen in training. Let $\tau$ and $\tau_{E}$ be a set of trajectories $(s_{1},a_{1},\cdots, s_{T},a_{T})$ generated by policy $\pi$ and expert policy $\pi_{E}$ repectively where each episode has a variable length $T$. We assume that there exists a sub-task $c_{t} \in \mathcal{C}$ at any given time step corresponding to each state-action pair $(s_{t}, a_{t})$, expert trajectories $\{\tau^{1}, \dots, \tau^{N}\} \in \tau_{E}$ are not labeled for the sub-tasks and each trajectory has a corresponding sequence of the sub-task variables $c = \{c_{1}, \cdots, c_{T}\}$.
\\ 

\begin{figure}[t]
    \centering
    \vspace{0.2cm}
    \includegraphics[angle=0, width=\columnwidth]{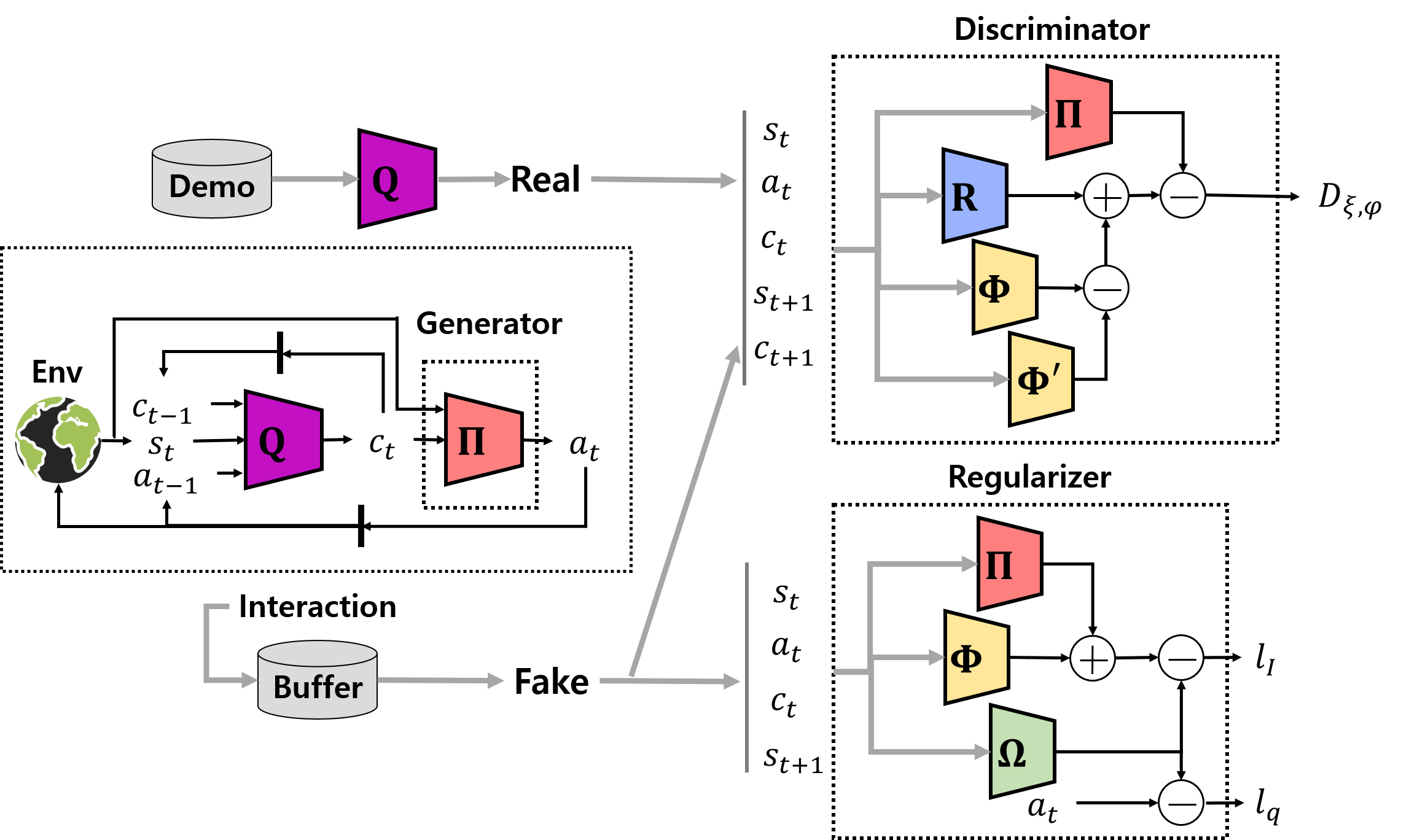}
    \caption{The overall adversarial architecture. A real sequence is obtained with pseudo-labels for the sub-tasks provided by the posterior. As the posterior and policy interact with the environment, a fake sequence is obtained. The discriminator models the reward function explicitly with the potential function that explains shaping term. The regularizer helps learned policy and reward function to be more robust. Subsequently, we use the discriminant signals $D_{\xi, \varphi}$ to train the hierarchical policy.}
    \label{fig:architecture}
\end{figure}

\begin{figure}[!t]
    \vspace{-0.5cm}
    \centering
    \begin{minipage}{\columnwidth}
\begin{algorithm}[H]
    \caption{Situational Empowerment-based Adversarial Inverse Reinforcement Learning}\label{alg:seairl}
    \begin{algorithmic}
    \State Obtain $\tau_E$ by running $\pi_E$ and $Q_{\psi}$
    \State Pretrain $Q_{\psi}$ using VAE with the gradient:
    \begin{equation}
        \resizebox{0.9\columnwidth}{!}{$
        - \mathbb{E}_{\tau_{E}} [\nabla_{\theta} \log \pi_{\theta}(a \vert s,c)] + \nabla_{\psi} D_{KL}[Q_{\psi}(c' \vert c, s', a) \, \vert \vert \, \mathcal{N}(0,I)] \nonumber
        $}
    \end{equation}
    \State Initialize parameters of $\pi_{\theta}$, $\Omega_{\phi}$, $\Phi_{\varphi}$ and $r_{\xi}$
    \State Synchronize the parameters of $\Phi_{\varphi'}$ with $\varphi$
    \For{$i \gets 0$ \textbf{to} $N$} 
        \State Collect trajectories $\tau$ by executing $\pi_{\theta}$ and $Q_{\psi}$
        
        \State Update $\phi_{i}$ with gradient $\mathbb{E}_{\tau}[\nabla_{\phi_{i}} l_{q}(s,a,c,s')]$
        
        \State Update $\varphi_{i}$ with gradient $\mathbb{E}_{\tau}[\nabla_{\varphi_{i}} l_{I}(s,a,c,s')]$
        
        \State Update $\xi_{i}$ with the gradient :
        \begin{equation}
            \resizebox{0.9\columnwidth}{!}{$
            \mathbb{E}_{\tau}[\nabla_{\xi_{i}} \log D_{\xi_{i}, \varphi_{i+1}}(s,a,c,s')] + \mathbb{E}_{\tau_{E}}[\nabla_{\xi_{i}}( 1 - \log D_{\xi_{i},
            \varphi_{i+1}}(s,a,c,s') )] \nonumber
            $}
        \end{equation}
        
        \State Update $\theta_{i}$ using PPO \cite{schulman2017proximal} with the gradient:
        \begin{equation}
            \resizebox{0.9\columnwidth}{!}{$
            \mathbb{E}_{\tau}[ \nabla_{\theta_{i}} \log \pi_{\theta_{i}}(a \vert s,c) \hat{r}_{\xi_{i+1}}(s,a,c,s',c')] - \lambda_{I} \mathbb{E}_{\tau}[\nabla_{\theta_{i}} l_{I}(s,a,c,s')] \nonumber
            $}
        \end{equation}
    \State After every n epoch synchronize $\varphi'$ with $\varphi$
    \EndFor
    \end{algorithmic}
\end{algorithm}
\end{minipage}
\vspace{-0.5cm}
\end{figure}

\begin{figure*}[!t]
    \centering
    \subfigure[MT10]{\includegraphics[width=0.46\linewidth]{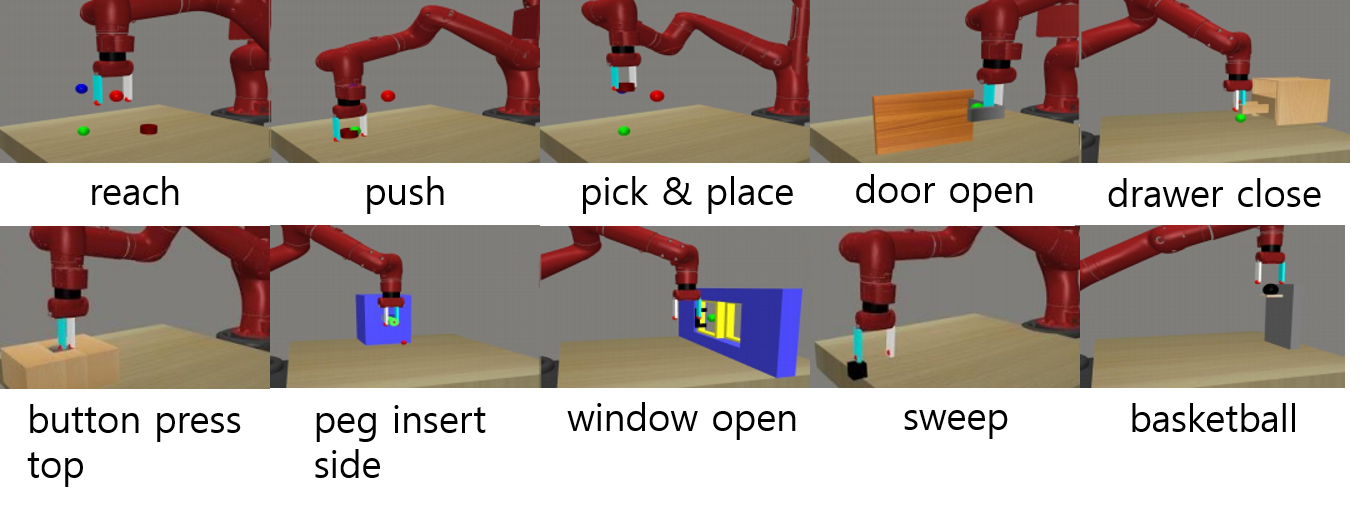}}
    \subfigure[ML5]{\includegraphics[width=0.46\linewidth]{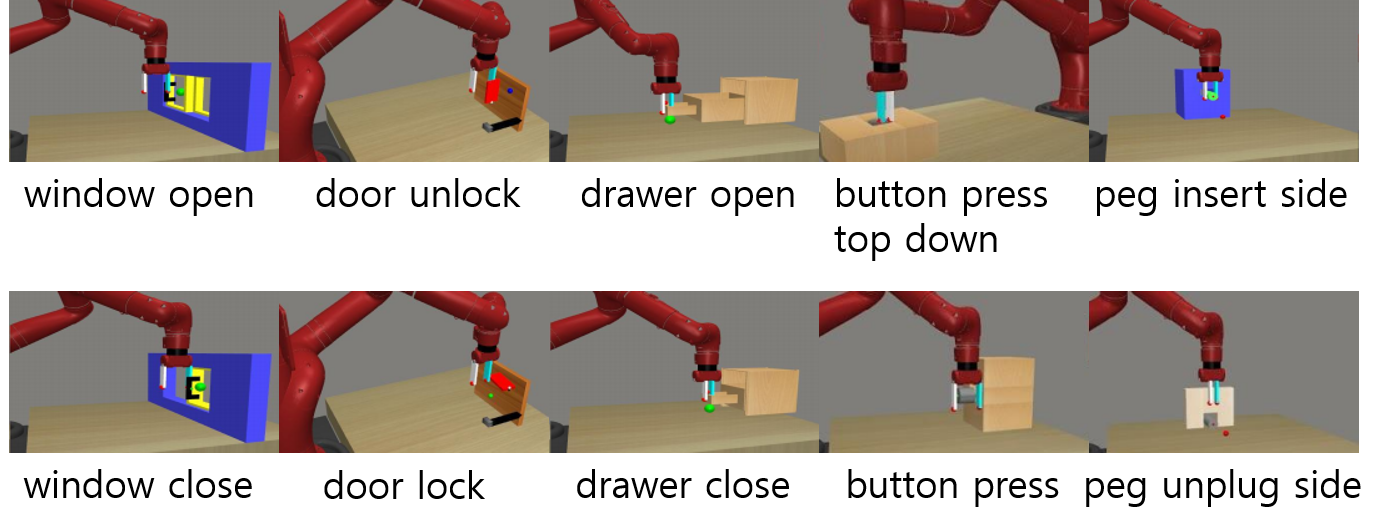}}
    \caption{Two environment settings for multi-task (a) and multi-task transfer (b) learning. \textbf{Left}: MT10 contains 10 random distinct scenarios and each of the scenarios had 50 randomized positions of the initial object and final goal. \textbf{Right} : The customized environment called ML5 (b) also has the equivalent setting with (a) for each scenario. In the phase of collecting expert demonstrations and pre-training with them, we randomly select one of the five scenarios in the first row. In the transfer learning phase, we choose one of the five scenarios in the second row.}
    \vspace{-0.5cm}
    \label{fig:metaworld}
\end{figure*}

\section{Proposed Method}

Our proposed method comprises five networks modeled as neural networks with each of the parameters, as shown in Fig. \ref{fig:networks}. (\lowercase\expandafter{\romannumeral1}) policy model outputs a distribution over actions given both the current state and the sub-task. (\lowercase\expandafter{\romannumeral2}) reward is a function of the state, action and sub-task. (\lowercase\expandafter{\romannumeral3}) a potential function determines the shaping term of reward $F=\gamma \Phi_{\varphi}(s',c') - \Phi_{\varphi}(s, c)$ and regularizes the policy to be updated. (\lowercase\expandafter{\romannumeral4}) an inverse model outputs a distribution over actions that bring about state transitions. (\lowercase\expandafter{\romannumeral5}) a posterior outputs a sub-task distribution given both the sub-tasks and the trajectory discovered up to the current time. The structures and hyperparameters of all networks are same with previous works \cite{sharma2018directed, qureshi2018adversarial}. All these models, except for the posterior, are trained simultaneously based on the objective functions described in Alg. \ref{alg:seairl}. The following sections explain how to learn the hierarchical policy and the multi-task reward function concurrently. Moreover, we reveal that our approach has the identical objective of the hindsight inference approach \cite{eysenbach2020rewriting}.

\subsection{Empowerment-Based Regularization for Multi-Tasking}

We introduce the sub-task variable to solve the complex multi-task learning problem with unknown dynamics. To provide robustness to environmental changes, we propose a method that recovers a multi-task reward function included in the discriminator, as shown in Fig. \ref{fig:architecture}. We define a situational empowerment $\Phi(s,c) = \underset{w, \Omega}{max} \; I^{w,\Omega}(a;s'|s,c)$ as a new internal reward. The variational lower bound, derived in Eq. \ref{eq:empoweredlowerbound} is expanded to the following inequality Eq. \ref{eq:situationalempoweredlowerbound}, and the detailed derivation is given in Appendix A.

\begin{align}
    &\mathbb{E}_{w}[-\log w(a|s,c)] 
    + \mathbb{E}_{Q,w,P}[\log \Omega(a|s,c,s')] \nonumber\\
    &= L_{I}(w,\Omega) \leq I(a;s'|s,c)
    \label{eq:situationalempoweredlowerbound}
\end{align}

We optimize the lower bound of $\hat{I}_{w,\Omega}$ over the distribution $w$ and hierarchical inverse model $\Omega$ using the expectation-maximization (EM) algorithm, where $w$ is the action distribution conditioned on the state and sub-task pair. For the hierarchical inverse model, we calculate the following objective Eq. \ref{eq:situationalinversemodelobejctive} based on the maximum log-likelihood problem:

\begin{align}
    l_{q}(s,a,c,s') = \vert \Omega_{\phi}(\hat{a}|s,c,s') - a \vert^{2} 
    \label{eq:situationalinversemodelobejctive}
\end{align}

For $w(a|s,c)$, which is the normalized hierarchical policy, we find the optimal solution $w^{*}(a|s,c) = e^{(\lambda - 1) + \beta \mathbb{E}_{Q,P} [\log \Omega (a|s,c,s')] }$ by replacing the constrained form $\frac{\partial \hat{I}^{w, \Omega}}{\partial w} = 0 \;s.t \sum_{a} w(a|s,c) = 1$ with an unconstrained Lagrange dual problem. We set the potential function $\Phi(s,c)$ as $\log Z(s,c)$ because $e^{(1 - \lambda)}$ denotes the partition function $Z(s,c) = \sum_{a}e^{\beta \mathbb{E}_{Q,P} [\log \Omega (a|s,c,s')]}$. 
We minimize the MSE between the approximated $w_{\theta,\varphi}(a|s,c) \approx \log \pi_{\theta}(a|s,c) + \Phi_{\varphi}(s,c)$ and $\log \Omega(a|s,c,s')$ as the follows: 


\begin{equation}
    \resizebox{0.9\columnwidth}{!}{$
    l_{I}(s,a,c,s') = \vert \log \Omega(a|s,c,s') 
    - \{\log \pi_{\theta}(a|s,c) + \Phi_{\varphi}(s,c)\} \vert^{2}
    $}
    \label{eq:situationalempowermentobjective}
\end{equation}

Finally, the objective derived in Eq. \ref{eq:inversemodelobjective} and Eq. \ref{eq:empowermentobjective} are extended into Eq. \ref{eq:situationalinversemodelobejctive} and Eq. \ref{eq:situationalempowermentobjective} by introducing latent code $c$ to deal with multi-task learning problems.

\subsection{Connection with Hindsight Relabeling Framework}

We design a reward function that represents the situational intention contained in expert distribution with an energy function based on sampling-based MaxEnt-IRL \cite{finn2016guided}. To extend the previous works \cite{fu2017learning, qureshi2018adversarial} to the multi-task learning, we propose a method to restore the multi-task reward function $f_{\xi, \varphi} (s,a,c,s',c') = r_{\xi}(s,a,c) + \gamma \Phi_{\varphi^{'}} (s', c') - \Phi_{\varphi}(s,c)$ that departs from changes in task dynamics by introducing a latent variable $c$. For learning stability, we fix the parameter of the target situational potential function $\Phi_{\varphi'}$ and update it at several intervals. With the reward function, we can obtain an optimal situational discriminator, as expressed by Eq. \ref{eq:situatonaldiscriminator}, where $z=f_{\xi, \phi}(s,a,c,s',c')-\log \pi(a \vert s,c)$.

\begin{align}
    \frac{e^{f_{\xi, \phi}(s,a,c,s',c')} }{e^{f_{\xi, \phi}(s,a,c,s',c')} + \pi(a|s,c)} 
    = \frac{1}{1+e^{-z}}
    \label{eq:situatonaldiscriminator}
\end{align}

When the above discriminator is plugged into Eq. \ref{eq:gailobjective}, we can obtain Eq. \ref{eq:seairlrewardobjective} for the reward model. Moreover, we can obtain Eq. \ref{eq:seairlpolicyobjective} for the policy model by adding posterior lower bound term in Eq. \ref{eq:lowerboundposterior} and empowerment-based regularization term in Eq. \ref{eq:situationalempowermentobjective}. 

\begin{figure*}[!t]
    \centering
    \subfigure[Hopper]{
        \includegraphics[width=.18\linewidth]{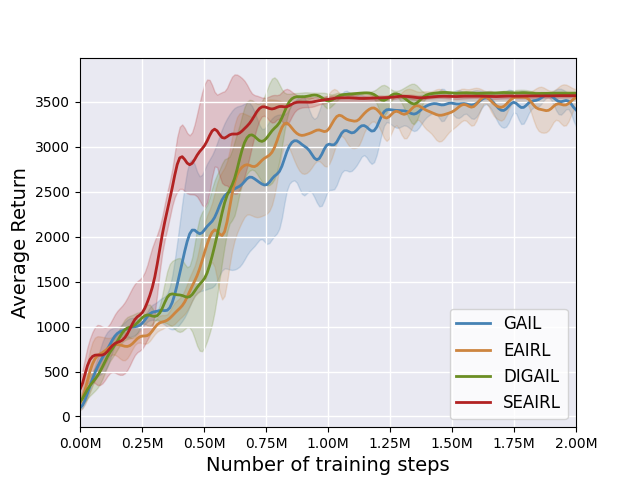}
        }
    \hspace{-0.2cm}
    \subfigure[Walker]{
        \includegraphics[width=.18\linewidth]{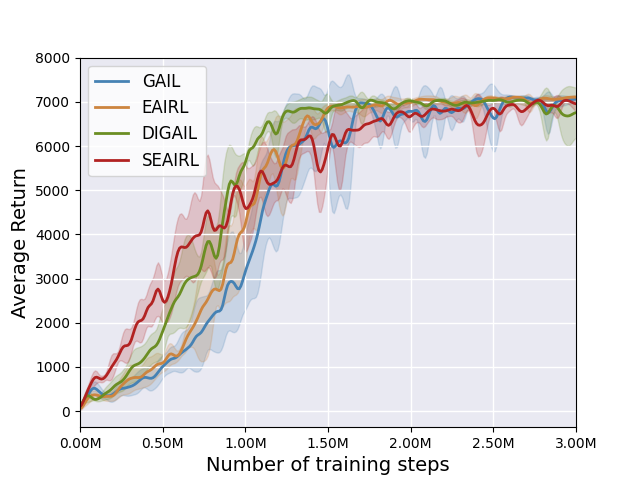}
        }
    \hspace{-0.2cm}
    \subfigure[Fetch]{
        \includegraphics[width=0.18\linewidth]{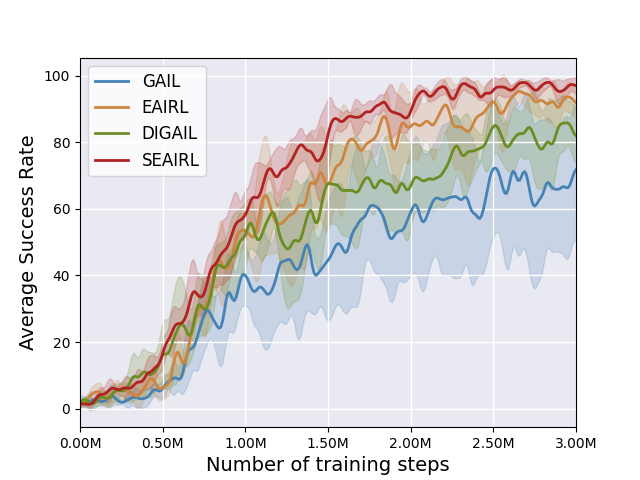}}
    \hspace{-0.2cm}
    \subfigure[MT10]{
        \includegraphics[width=.18\linewidth]{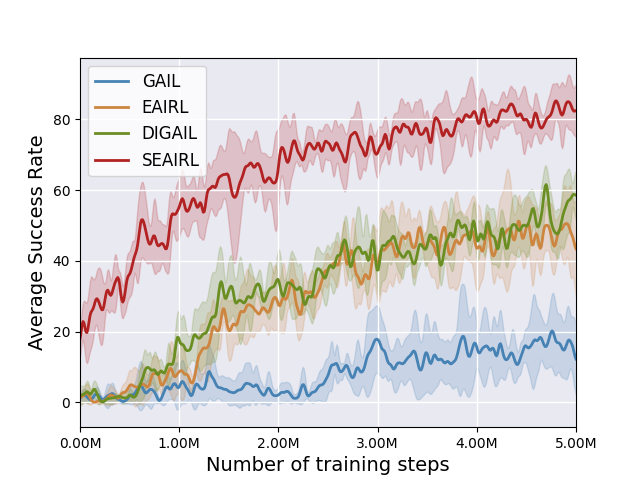}
        }
    \hspace{-0.2cm}
    \subfigure[ML5]{
        \includegraphics[width=.18\linewidth]{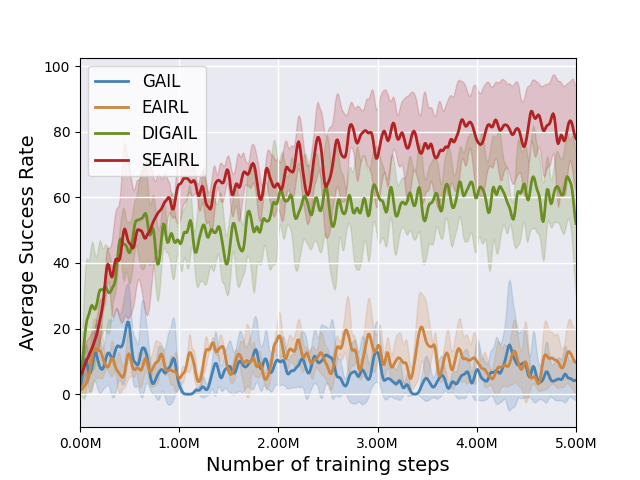}
        }
    \caption{Learning curves for Hopper, Walker, Fetch, MT10, and ML5 environments. The darker-colored lines and shaded areas denote the average return (a, b) or success rate (c-e) and standard deviations respectively, computed over 5 random seeds.}
    \vspace{-0.2cm}
    \label{fig:learning_curve}
\end{figure*}

\begin{align}
    &\max_{\xi} \mathbb{E}_{\tau}[ \log D_{\xi,\varphi}(s,a,c,s',c') ] \nonumber\\
    &\;\;\;\;\;\;\;\;\;\;+ \mathbb{E}_{\tau_{E}} [\log (1 - D_{\xi,\varphi}(s,a,c,s',c'))] 
    \label{eq:seairlrewardobjective}
    \\
    &\min_{\theta} \mathbb{E}_{Q,\pi}  [- f_{\xi, \varphi} (s,a,c,s',c') + \log \pi_{\theta}(a|s,c)] \nonumber\\
    &\;\;\;\;\;\;\;\;\;\;+\lambda_1 L_{q}(\pi,Q) + \lambda_2  l_{I} (\pi, \Phi)
    \label{eq:seairlpolicyobjective}
\end{align}

To apply the policy gradient, we can rewrite Eq. \ref{eq:seairlpolicyobjective} into Eq. \ref{eq:alternativeWay} by replacing it with the alternative reward $\hat{r}_{\xi}(s_{t},a_t,c_{t},s_{t+1},c_{t+1}) = f_{\xi, \varphi} (s_{t},a_t,c_{t},s_{t+1},c_{t+1}) + \lambda_{q} \log  Q_{\psi}(c_{t}\vert s_t, a_{t-1}, c_{t-1})$.

\begin{align}
    &\max_{\theta} \mathbb{E}_{\tau} [\log \pi_{\theta}(a_t|s_t,c_t) \hat{r}_{\xi}(s_{t},a_t,c_{t},s_{t+1},c_{t+1})] \nonumber \\
    &\;\;\;\;\;\;\;\;\;\;+\lambda_{h}H(\pi_{\theta}) - \lambda_{I} l_{I} (\pi, \Phi)
    \label{eq:alternativeWay}
\end{align}

Therefore, our proposed method is a special case of the previous work Eq. \ref{eq:mutitaskobjective} because maximizing $\log  Q_{\psi}(c_{t}\vert s_t, a_{t-1}, c_{t-1})$ is the same as minimizing $D_{KL}(Q(c,|s,a) || p(c))$.

\begin{table}[!t]
\centering
    \caption{Comparison of Average returns with standard deviation}
    \begin{adjustbox}{max width=0.9\columnwidth}
        \begin{tabular}{c c c c}
            \toprule
            \textsc{Method} & \textsc{Hopper} & \textsc{Walker} & \textsc{Fetch} \\
            \midrule
            GAIL & 3604.94 $\pm$ 18.85 & 7128.18 $\pm$ 710.64 & -7.65 $\pm$ 5.15 \\
            
            DIGAIL & \textbf{3632.48} $\pm$ 9.37 & 7262.67 $\pm$ 138.09 & -6.13 $\pm$ 4.99 \\
            
            EAIRL & 3615.72 $\pm$ 7.54 & \textbf{7339.17 $\pm$ 41.24} & -3.67 $\pm$ 2.07 \\
            
            SEAIRL &  3630.86 $\pm$ \textbf{5.69}  & 7212.81 $\pm$ 49.92 & \textbf{-3.31 $\pm$ 1.78} \\
            \bottomrule
        \end{tabular}
    \end{adjustbox}
    \vspace{-0.5cm}
    \label{tab:mujoco_result}
\end{table}

\section{Experiments}

We present the experimental results of both multi-task and multi-task transfer learning by utilizing physics engines (i.e., MuJoCo / Robotics / Meta-World) interfaced within OpenAI Gym \cite{brockman2016openai, yu2020meta}. We validate the outstanding performance of our proposed method (SEAIRL) compared with existing baselines (i.e., GAIL / DIGAIL / EAIRL). In the multi-task learning, we show that the more complex the relation of the task transition, the better the performance of our proposed model. Furthermore, we demonstrate robustness through successful transfer learning in environments with work shifts that are not seen when acquiring expert trajectories. 

\begin{figure}[!th]
    \centering
    \vspace{-0.3cm}
    \subfigure[MT10]{
        \includegraphics[width=1.0\columnwidth]{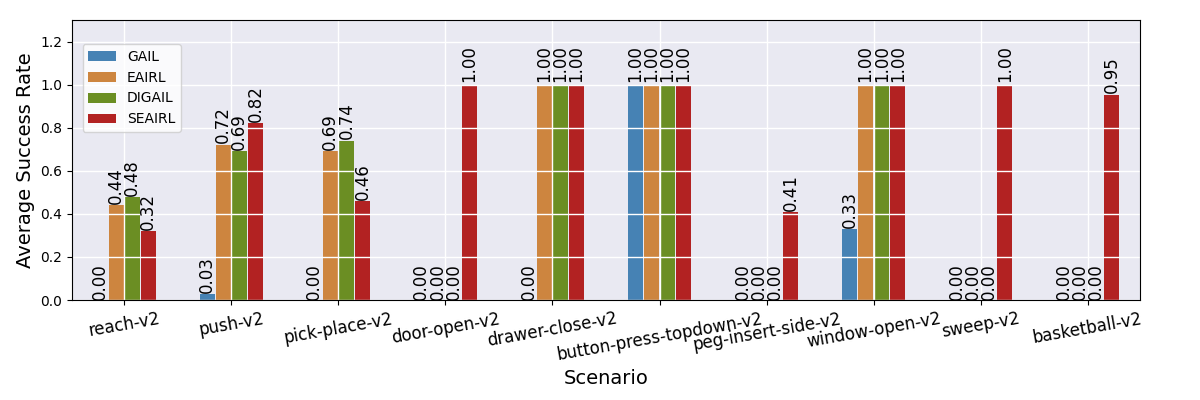}
        }
    \subfigure[ML5]{
        \includegraphics[width=1.0\columnwidth]{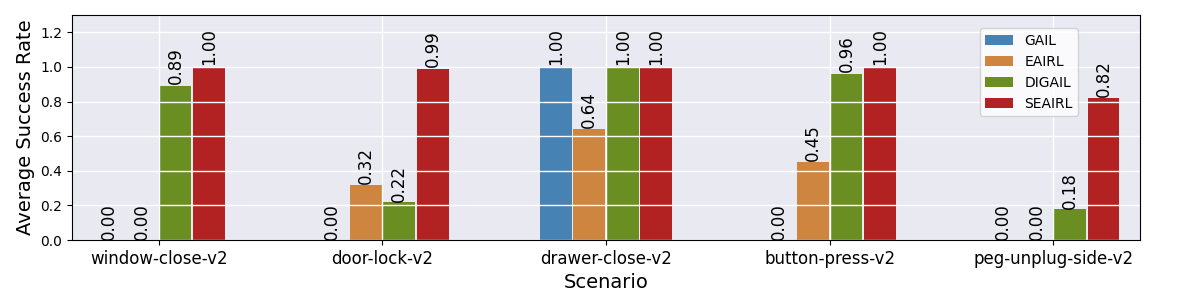}
        }
    \caption{The average success rate for MT10 (a) and ML5 (b) environment. The average rate of each scenario is computed over 100 episodes.}
    \vspace{-0.3cm}
    \label{fig:scenario_success_rate}
\end{figure}

\subsection{Multi-Task Learning Performance}

To validate that our proposed approach could acquire distinct skills in multiple tasks, we experimented with continuous state-action control tasks. Hopper, Walker, and FetchPickandPlace (Fetch) are single scenarios composed of a few sub-tasks. Specifically, for Hopper, macro actions, such as jumping, mid-air, and landing, are switched periodically. In case of Walker, the agent lifts and sets down each foot in turn at a regular pace. In Fetch, the agent controls the end effector to grasp and lift a block, and then reach a goal point where the positions of the block and goal are randomly initialized. Fig. \ref{fig:learning_curve}(a-c) shows the learning curve of each environment in a single scenario and Table. \ref{tab:mujoco_result} represents the final average return with standard deviations for each method over 100 episodes. Overall, the curves show that our method has better data efficiency. In contrast to Walker and Hopper, our proposed model exhibits better performance in Fetch with randomness. Meanwhile, for the case in which the dimension increases from Hopper to Walker, the proposed model shows an indistinctive final performance over the baselines.

\begin{table}[!t]
\centering
\vspace{-0.2cm}
\caption{Comparison of Average Success Rate}
    \begin{tabular}{c c c c}
        \toprule
        \textsc{Method} & \textsc{Fetch} & \textsc{MT10} & \textsc{ML5} \\
        \midrule 
        GAIL & 0.46 & 0.14 & 0.20 \\
        
        DIGAIL & 0.73 & 0.46 & 0.72 \\
        
        EAIRL & 0.82 & 0.43 & 0.30 \\
        
        SEAIRL & \textbf{0.97} & \textbf{0.82} & \textbf{0.98} \\
        \bottomrule
    \end{tabular}
    \vspace{-0.5cm}
    \label{tab:success_rate_result}
\end{table}

\begin{figure}[!ht]
    \centering
    \vspace{-0.1cm}
    \subfigure[Unnormalized Reward]{
        \includegraphics[width=.46\columnwidth]{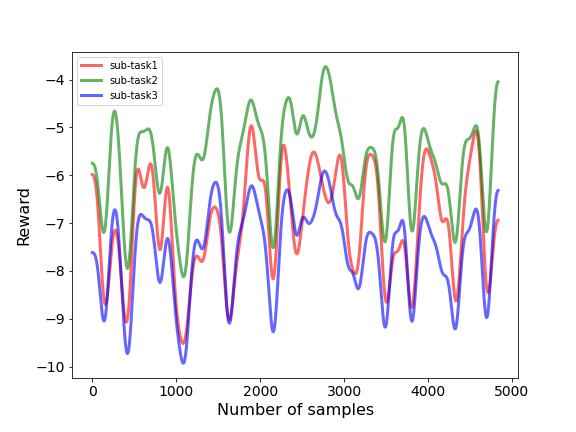}
        }
    \subfigure[Normalized Reward]{
        \includegraphics[width=.46\columnwidth]{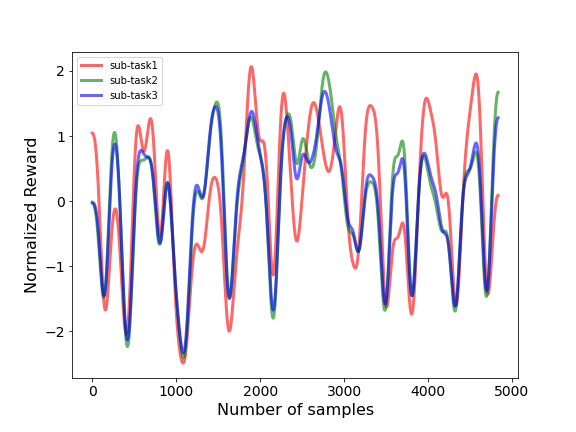}
        }
    \caption{The effect of situational empowerment that normalizes rewards with different scales for each sub-task where the sub-task variable size is 3. The samples are obtained in MT10 environment. Unlike the Unnormalized rewards (a), the normalized rewards (b) are comparable for each sub-task.}
    \vspace{-0.2cm}
    \label{fig:normalized_reward}
\end{figure}

For comparison in environments with more complex task, we carried out the test in the MT10 environment as shown in Fig. \ref{fig:metaworld}(a). We expect that the reward function trained with our method provide meaningful learning signals to learn a composable sub-task policy. Subsequently, the agent can successfully perform tasks in various scenarios by combining sub-tasks. Compared with the baselines, our approach exhibits a significant improvement in performance and convergence speed as shown in Fig. \ref{fig:learning_curve}(d). The final average success rate over randomly selected 100 episodes for each method is shown in Table. \ref{tab:success_rate_result}. Our approach exhibits 1.18 and 1.78 times higher performance than the existing methods in Fetch and MT10 environments respectively, which means that our method performs better as the randomness and task complexity increase. Fig. \ref{fig:scenario_success_rate}(a) shows the average success rate for each scenario. GAIL tends to overfit only in one scenario. EAIRL also has a limitation in targeting multiple tasks. In case of DIGAIL, it can be difficult for the policy to understand the hierarchical structure unless the reward signal is explicitly expressed for each task. Our method compensates for the shortcomings of the two methods by normalizing the reward scales for each task, which provides reward signals to the hierarchical policy to explore a meaningful futures, as shown in Fig. \ref{fig:normalized_reward}. By doing so, we successfully solved many scenarios that have zero rates in the baselines. As a result, we validated a strong point of multi-task learning.

\begin{figure}[!t]
    \centering
    \vspace{-0.1cm}
    \adjustbox{max width=\columnwidth}{
    \subfigure[Labels with scenario]{
        \includegraphics[width=.46\columnwidth]{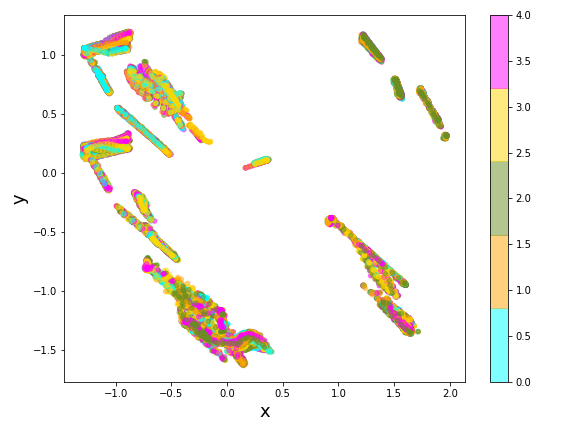}
        }
    \subfigure[Labels with sub-task]{
        \includegraphics[width=.46\columnwidth]{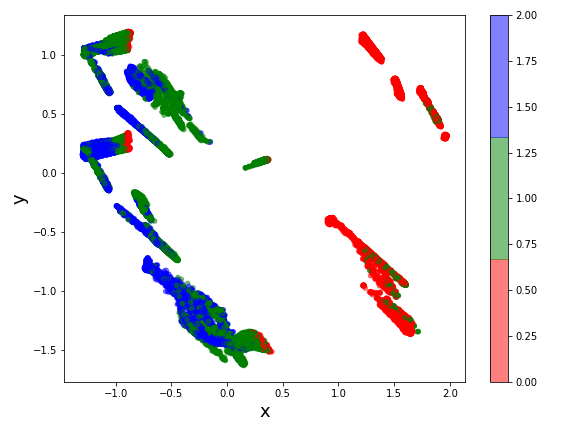}
        }
    }
    \caption{Visualization of sub-tasks shared in different scenarios of ML5 environment. Unlike labeling with scenarios (a), we can cluster state-action samples samples by classifying them with sub-tasks like (b).}\label{fig:sub_task_vis}
    \vspace{-0.5cm}
\end{figure}

\subsection{Multi-Task Transfer Learning Performance}

\begin{figure}[!t]
    \centering
    \vspace{-0.1cm}
    \subfigure[door unlock]{\includegraphics[width=0.46\columnwidth]{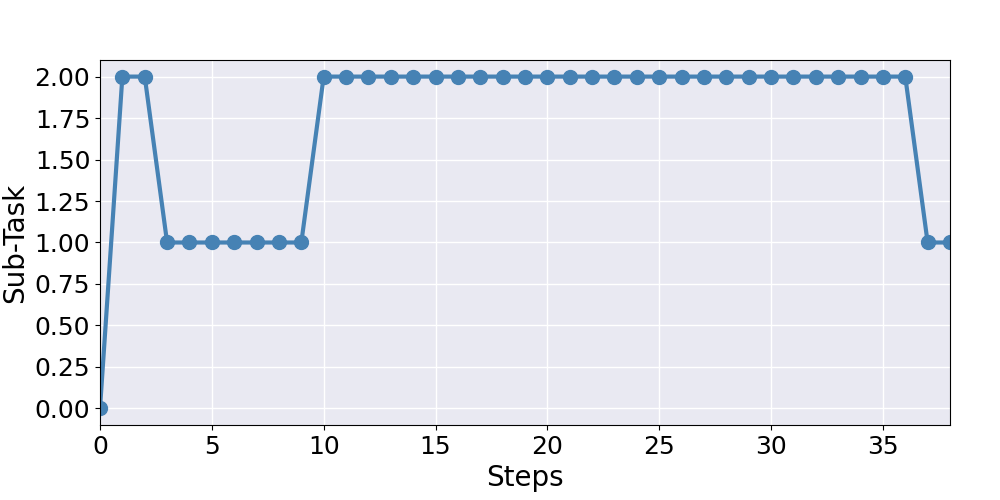}
    }
    \subfigure[button press top down]{\includegraphics[width=0.46\columnwidth]{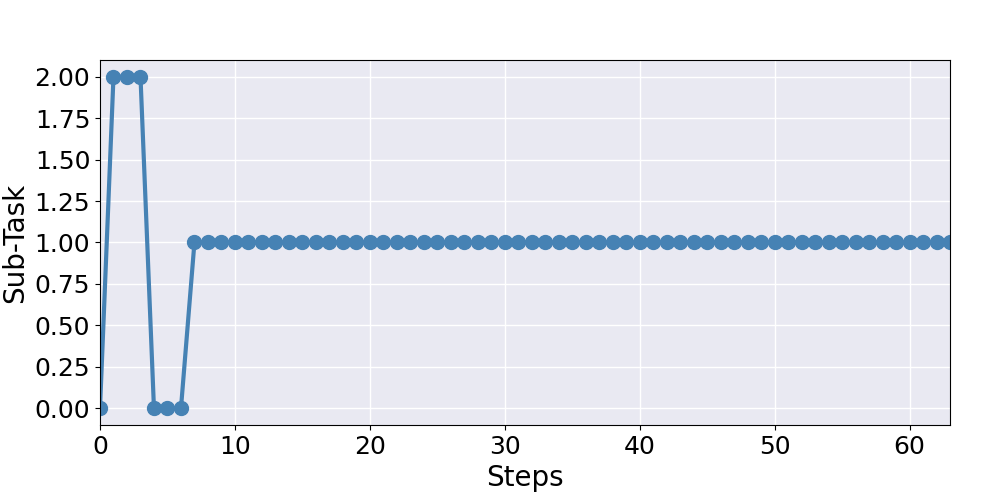}
    }
    \subfigure[door lock]{\includegraphics[width=0.46\columnwidth]{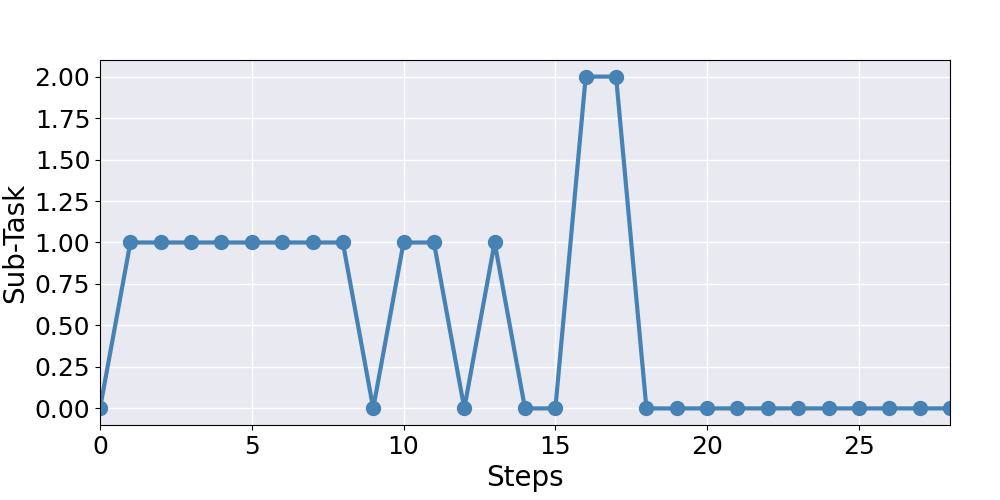}
    }
    \subfigure[button press]{\includegraphics[width=0.46\columnwidth]{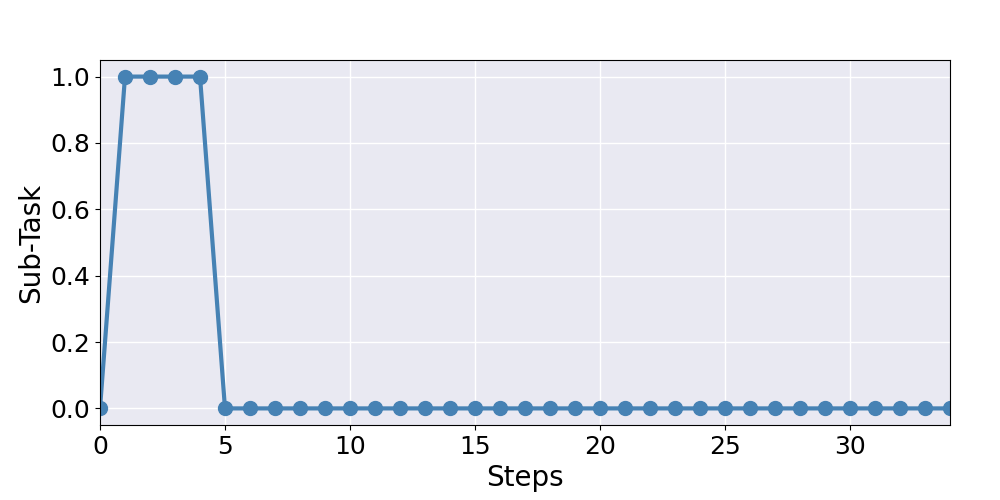}
    }
    \caption{The visualization of sub-task transition to adapt to the new environments. The sub-task is a 3 dimensional one-hot vector represented by a categorical variable with Gumbel-softmax trick \cite{jang2016categorical}.}
    \vspace{-0.5cm}
    \label{fig:adapt_visualization}
\end{figure}

To confirm the ability to quickly adapt to new multitasks, we built an ML5 environment as shown in Fig. \ref{fig:metaworld}(b). To solve the different scenarios, we find common sub-tasks shared by each scenario through the posterior as shown in Fig. \ref{fig:sub_task_vis}. We expect that learned rewards disentangled from sub-tasks can be composed to give learning signals that guide policy into new types of behavior. In turn, the policy chains learned macro actions together to create a different desired policy, as shown in Fig. \ref{fig:adapt_visualization}. The swapping pattern of the latent variable used in the environments in the first row is recombined according to the changed ones in the second row. 

Our approach outperforms both GAIL and EAIRL as well as DIGAIL in terms of data efficiency and performance, as illustrated in Fig. \ref{fig:learning_curve}(e). This is because both GAIL and EAIRL do not focus on multi-tasking and DIGAIL does not model the multi-task reward function explicitly. Fig. \ref{fig:scenario_success_rate}(b) represents the success rate of the newly adapted scenario through transfer learning. GAIL shows a tendency to overfit only in one scenario, similar to the MT10 environment. In the case of EAIRL, even though generality has increased in several scenarios, but the overall performance has decreased. It is difficult for DIGAIL to adapt to a new task. According to Table. \ref{tab:success_rate_result}, the proposed method results in a state-of-the-art average success rate over 100 episodes by 1.36 times compared to the existing method in the ML5 environment. Finally, we verified the robustness of changing task dynamics in multiple scenarios.

\section{CONCLUSIONS}

Our work proposes an approach to recover transferable rewards and obtain an adaptable policy through adversarial learning using unstructured demonstrations. The key idea is to learn situational empowerment through the maximization of conditional mutual information on sub-tasks. We also show the theoretical connections with the hindsight inference literature. The proposed regularization normalizes reward signals for each task, which in turn prevents the hierarchical policy from overfitting into local behavior. By outperforming baselines, we show that our policy can handle highly diverse tasks and adapt quickly to dynamically different environments. We demonstrate that our learned reward and policy understand hierarchical task structure without predefined knowledge and lead to meaningful generalization across many tasks and unknown dynamics.

In the future, we will extend our work to learn a posterior model concurrently based on graphical embedding to naturally deal with long-term or image-based multi-tasks. Another exciting direction would be to investigate how to build an algorithm that learns from suboptimal demonstrations that contain both optimal and non-optimal behaviors.


\section*{APPENDIX}
\subsection{Variational Information Lower bound}
\label{appendix:variationallowerbound}

By defining Mutual Information (MI) as a difference in conditional entropies $I^{w, \Omega}(a ; s' ;\vert s, c) = H(a \vert s, c) - H(a \vert s, c, s')$ as mentioned in section \uppercase\expandafter{\romannumeral4}-A, the variational lower bound representation of MI is derived as follow:

\begin{align}
    & -  H(a_{t}  \vert s_{t}, c_{t}, s_{t+1}) \nonumber \\
    &= \mathbb{E}_{\substack{Q(c_{t} \vert s_{t}, a_{t-1}, c_{t-1}) \\ w(a_{t} \vert s_{t}, c_{t}) \\ P(s_{t+1} \vert a_{t}, s_{t}) } }[\log p(a_{t} \vert s_{t}, c_{t}, s_{t+1})] \nonumber  \\
    &= \mathbb{E}_{Q,w,P}[\log \frac{p(a_{t} \vert s_{t}, c_{t}, s_{t+1}) \Omega(a_{t} \vert s_{t}, c_{t}, s_{t+1})}{\Omega(a_{t} \vert s_{t}, c_{t}, s_{t+1})}] \nonumber \\
    &= \mathbb{E}_{Q,w,P}[\Omega(a_{t} \vert s_{t}, c_{t}, s_{t+1}) + \log \frac{p(a_{t} \vert s_{t}, c_{t}, s_{t+1})}{\Omega(a_{t} \vert s_{t}, c_{t}, s_{t+1})}] \nonumber \\
    &= \mathbb{E}_{Q,w,P}[\Omega(a_{t} \vert s_{t}, c_{t}, s_{t+1})] + D_{KL} (p \,\vert\vert\, \Omega) \nonumber \\
    &\geq \mathbb{E}_{Q,w,P}[\Omega(a_{t} \vert s_{t}, c_{t},  s_{t+1})] \nonumber
\end{align}

\section*{ACKNOWLEDGMENT}
This work was supported by the Basic Science Research Program through the National Research Foundation of Korea (NRF) funded by the Ministry of Science and ICT(2017R1E1A1A01075171) and in part by the Institute of New Media and Communications and the Automation and Systems Research Institute, Seoul National University.


\bibliography{ref}
\bibliographystyle{ieeetr}

\end{document}